\documentclass[nohyperref]{article}

\usepackage{microtype}
\usepackage{graphicx}
\usepackage{subfigure}
\usepackage{booktabs} 

\usepackage[accepted]{icml2022}

\usepackage{amsmath}
\usepackage{amssymb}
\usepackage{mathtools}
\usepackage{amsthm}
\usepackage{xcolor}         
\usepackage{soul}
\usepackage[hang,flushmargin]{footmisc}
\usepackage{physics}
\usepackage{mathtools}
\usepackage{wrapfig}

\usepackage[capitalize,noabbrev]{cleveref}

\theoremstyle{plain}

\theoremstyle{definition}

\theoremstyle{remark}

\usepackage[textsize=tiny]{todonotes}

\icmltitlerunning{}

\begin{document}

\onecolumn
\icmltitle{AdamNODEs: When Neural ODE Meets Adaptive Moment Estimation}




\begin{icmlauthorlist}
\icmlauthor{Seunghyeon Cho}{yonsei}
\icmlauthor{Sanghyun Hong}{osu}
\icmlauthor{Kookjin Lee}{asu}
\icmlauthor{Noseong Park}{yonsei}
\end{icmlauthorlist}

\icmlaffiliation{yonsei}{Yonsei University}
\icmlaffiliation{osu}{Oregon State University}
\icmlaffiliation{asu}{Arizona State University}

\icmlcorrespondingauthor{Noseong Park}{noseong@yonsei.ac.kr}

\icmlkeywords{Machine Learning, ICML}

\vskip 0.3in



\printAffiliationsAndNotice{}  


\newcommand{\solution}{AdamNODEs}

\newcommand{\topic}[1]{\noindent \textbf{#1}}

\newcommand{\SH}[1]{\noindent {\color{red} \textbf{Sanghyun:} #1}}

%
\begin{abstract}

Recent work by Xia~\textit{et al.} leveraged the continuous-limit of the classical momentum accelerated gradient descent and proposed heavy-ball neural ODEs.
While this model offers computational efficiency and high utility over vanilla neural ODEs, this approach often causes the \emph{overshooting} of internal dynamics, leading to unstable training of a model.
Prior work addresses this issue by using ad-hoc approaches, \emph{e.g.}, bounding the internal dynamics using specific activation functions, but the resulting models do not satisfy 
the exact heavy-ball ODE.
In this work, we propose adaptive momentum estimation neural ODEs (\solution{}) that \emph{adaptively} control the acceleration of the classical momentum-based approach.
We find that its adjoint states also satisfy AdamODE and do not require ad-hoc solutions that the prior work employs.
In evaluation, we show that \solution{} achieve the lowest training loss and efficacy over existing neural ODEs.
We also show that \solution{} have better training stability than classical momentum-based neural ODEs.
This result sheds some light on adapting the techniques proposed in the optimization community to improving the training and inference of neural ODEs further.
Our code is available at {\footnotesize\url{https://github.com/pmcsh04/AdamNODE}}.
\end{abstract}

%
\section{Introduction}
\label{sec:intro}

Neural ordinary differential equations (NODEs)~\citep{chen2018neural} are continuous-depth neural networks that perform forward and backward passes by solving an ODE and its adjoint form.
NODEs model underlying continuous-time dynamics as a form of ODEs:
$\dv{\mathbf{h}}{t}\!=\!\mathbf{f}(\mathbf{h}(t), t;\theta_f)$,
where $\mathbf{f}$ is parameterized by a neural network with learnable parameters $\theta_f$.
The forward pass is equivalent to solving an initial value problem (IVP) with $\mathbf{f}$ and an initial condition $\mathbf{h}(t_0)$, and the backward pass solves its reverse-mode IVP.
By formulating the forward and backward passes into solving an ODE, 
NODEs improve memory efficiency and scalability of continuous-depth models. (see \S\ref{sec:related} for the details). 

\begin{wrapfigure}{r}{9.cm}
\centering
\vspace{-1.4em}     
\includegraphics[width=0.494\linewidth]{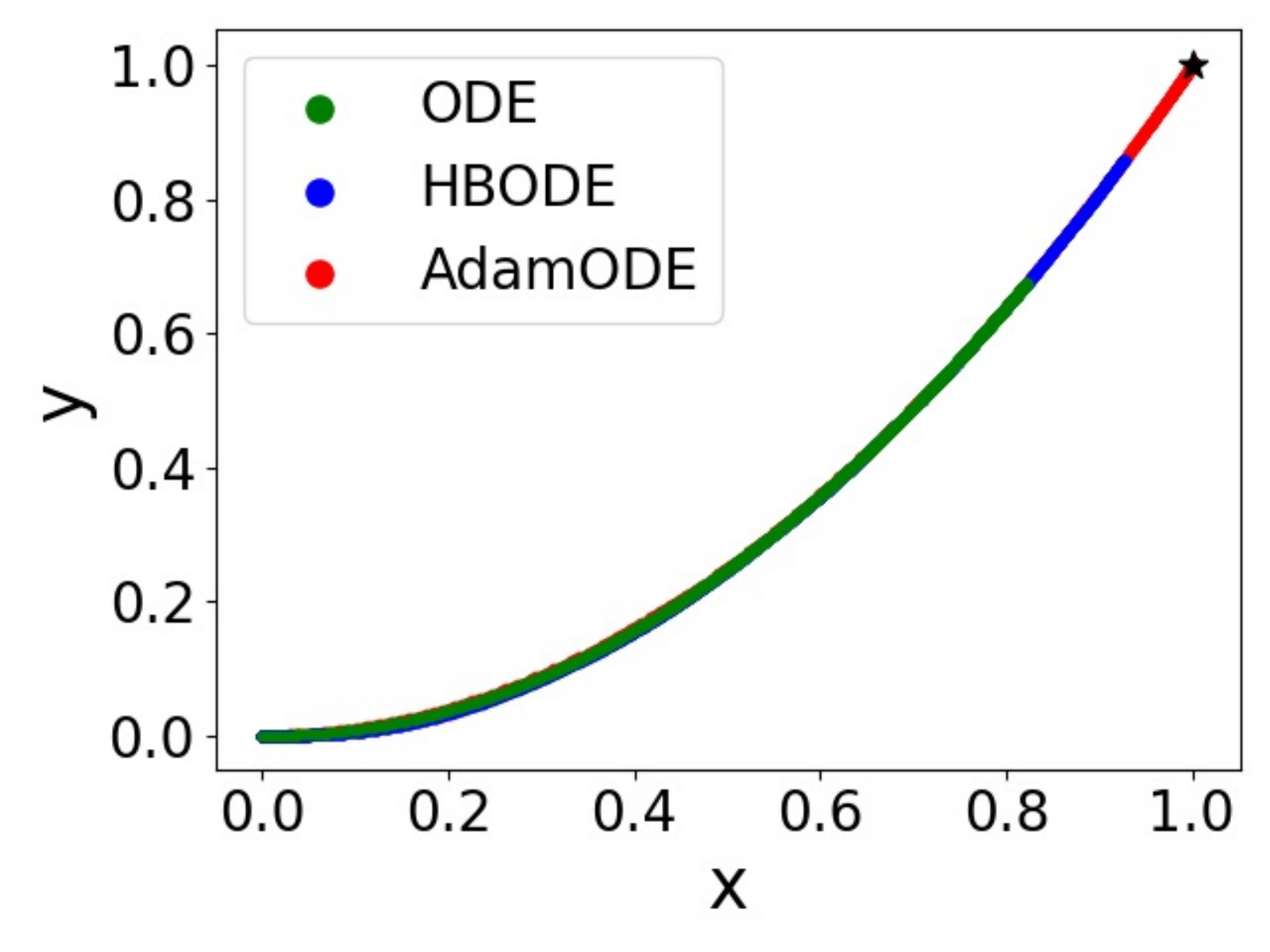}
\includegraphics[width=0.494\linewidth]{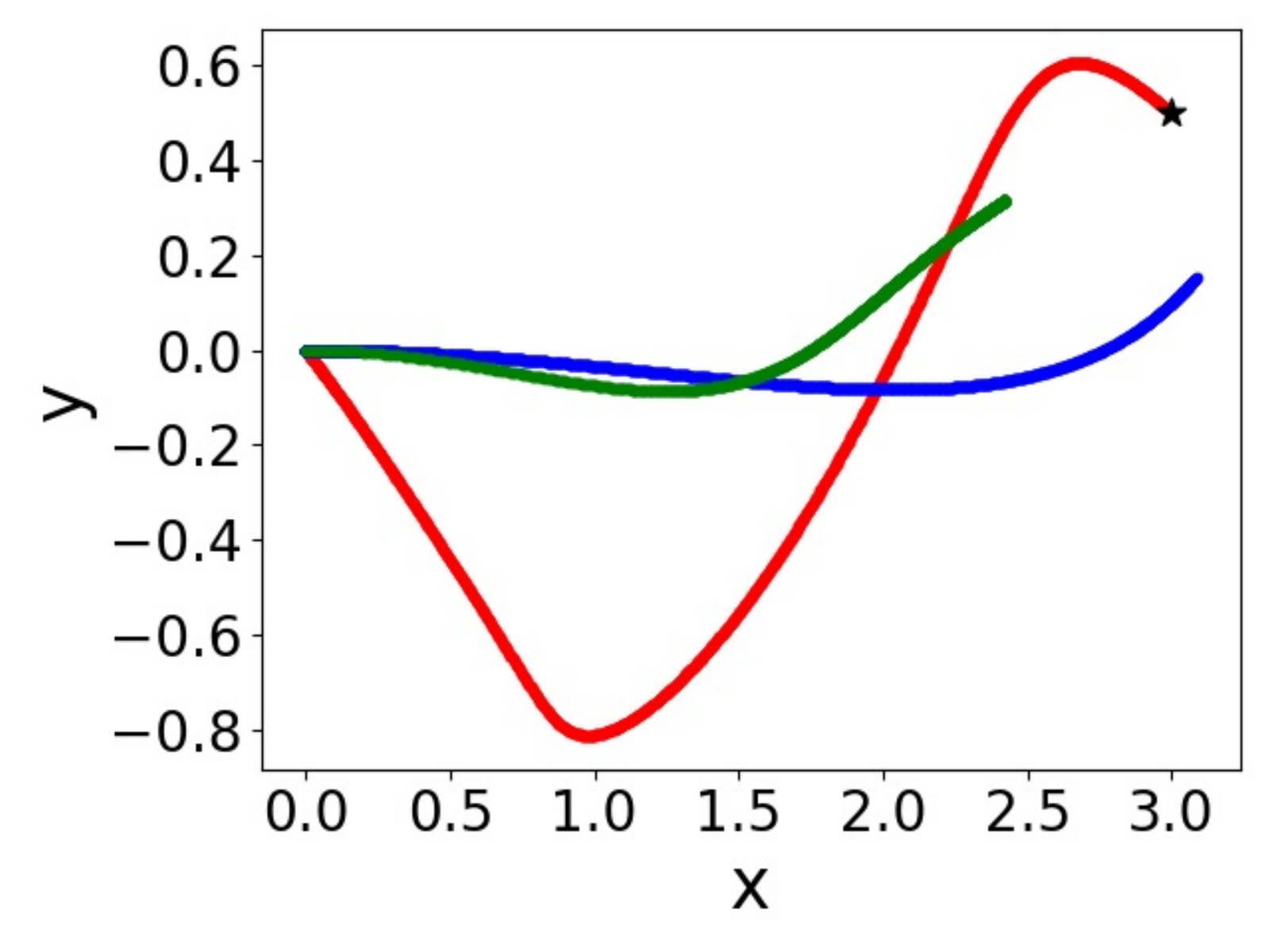}
\vspace{-2.4em}
 
\caption{\textbf{Trajectories computed by solving ODE, HBODE, and AdamODE.} We test three different ODE formulations on the Rosenbrock (left) and Beale (right) functions. AdamODE converges to the stationary point (marked as a star) closer and faster than the others.}
\label{fig:motivation}
\vspace{-1.em}
\end{wrapfigure}
Modeling underlying dynamics using the first-order equation, however, limits the expressivity of a model and often leads to sub-optimal performances. 
%
Prior work addresses this issue with diverse approaches.
\citet{dupont2019augmented} proposed Augmented Neural ODEs (ANODEs) that solve the ODE on the space with additional dimensions.
\citet{norcliffe2020second} proposed Second-Order Neural ODEs (SONODEs) and showed that we could generalize the adjoint sensitivity method in the first-order ODEs and solve SONODEs more efficiently.
Recent work by~\citet{xia2021heavy} leveraged those findings and proposed 
heavy-ball neural ODEs (HBNODEs) that model the dynamics with the continuous limit of the classical momentum method, suggesting that we may use the knowledge from the optimization literature to improve the training and inference of NODEs further.

In this work, we continue this research direction and characterize the benefit of using adaptive moment estimation in modeling NODEs.
Specifically, we ask the question: \emph{How can we model NODEs with adaptive moment estimation, and what are the benefits over existing NODEs?}
To motivate, we solve the ODEs with the vanilla, classical momentum (HBODE), and adaptive moment estimation (AdamODE) on the Rosenbrock and Beale functions 
and contrast the trajectories.
In Figure~\ref{fig:motivation}, we show that AdamODE converges to the stationary point 
faster and closer than ODE and HBODE and, thus, 
is efficient, stable, and accurate.
However, it is unknown whether these potential benefits can transfer to NODEs.

\topic{Contributions.}
\emph{First}, we propose \solution{} that model the underlying dynamics leveraging 
the continuous-limit of the adaptive estimates of lower-order moments.
Starting from the update rules of Adam optimizer~\citep{kingma2014adam}, we derive the adjoint form of \solution{}.
This adjoint form is composed of three adjoint equations, making it computationally intractable to solve them directly using the adjoint sensitivity method. 
\emph{Second}, we solve this form by leveraging the idea of augmenting the solution space, proposed by~\citet{dupont2019augmented}, and by projecting each equation to a different dimension.
\emph{Third}, we evaluate \solution{} on the two image classification tasks and the Silverbox task.
In two image classification tasks, \solution{} achieve the highest test accuracy per forward or backward NFE compared to the baselines.
In the Silverbox task, we show that \solution{} can mitigate the overshooting (or the blow-up) problem.

%
\section{Related Work}
\label{sec:related}

\topic{Continuous-depth neural networks} enable flexibility in modeling dynamics and have been studied as a potential alternative to traditional deep feed-forward neural networks~\citep{weinan2017proposal, haber2017stable, lu2018beyond}.
\citet{chen2018neural} addressed the main technical challenge of using continuous-depth networks in practice, \textit{i.e.}, efficient back-propagation through the ODE solver, by using the adjoint sensitivity method. 
Many continuous-depth models that share the same computational formalism 
have been proposed, \textit{e.g.}, Neural-Controlled DEs~\citep{kidger2020neural} and Neural Stochastic DEs~\citep{liu2019neural}.
Followup work used those models and has shown success in diverse tasks, such as image classification~\citep{zhuang2020adaptive, zhuang2021mali} 
or data-driven modeling of physical quantities~\citep{Greydanus2019hnn, cranmer2020lagrangian, lee2021machine}.
%
We propose a new continuous-depth network that leverages the adaptive moment estimation.

\topic{Modeling ODEs using second-order acceleration.}
Prior work
~\citep{polyak1964some, nesterov1983method, su2014differential, chen2014stochastic, wilson2021lyapunov} has shown that second-order ODEs with a damping term, such as classical or Nesterov momentum, accelerate the convergence of the first-order ODEs.
\citet{xia2021heavy} leverage this insight and model NODEs with the classical momentum method.
The adjoint state of the resulting HBNODEs satisfies an HBNODE, which significantly reduces the number of function evaluations (NFEs) required for forward/backward pass and improves the generalization.
However, the classical momentum often causes excessive acceleration, causing the \emph{overshooting} problem~\citep{norcliffe2020second}.
To address this issue, they propose Generalized HBNODEs (GHBNODEs) that bound the increase of second-order dynamics by using a nonlinear activation, \textit{e.g.}, hardtanh.
In contrast, we use adaptive moment estimation and mitigate the overshooting problem while achieving a better test accuracy and computational efficiency than HBNODEs and GHBNODEs.

%
\section{Adaptive Moment Estimation Neural ODEs}


\subsection{Preliminaries on Heavy-Ball Neural ODEs}
\label{subsec:heavy-ball}
%
NODEs, especially in high accuracy settings, require large computations in training and inference.
To mitigate this issue,~\citet{xia2021heavy} have proposed HBNODEs that leverage the concept of the classical momentum.
%
%
%
We first review the derivation of HBNODEs from the classical momentum equation (Eq.~\eqref{eq:classical_momentum}), where $\mathbf{F}$ is a generic function we optimize:
\begin{equation}\label{eq:classical_momentum}
    \mathbf{x}^{k+1} = \mathbf{x}^{k} -s\bigtriangledown \mathbf{F}(\mathbf{x}^{k}) + \beta (\mathbf{x}^{k}-\mathbf{x}^{k-1}),
\end{equation}
where $s>0$ is the step size and $0\leq\beta\leq1$ is the momentum scaler.
Let $m_k := (x_{k+1} - x_k)/\sqrt{s}$ and $\beta := 1 - \gamma \sqrt{s}$, where $\gamma\geq0$ is another hyperparameter.
Suppose we take the continuous-limit of this form (\textit{i.e.}, $s\rightarrow0$), Eq.~\eqref{eq:classical_momentum} becomes:
\begin{equation}\label{eq:HBODE}
    \frac{\partial \mathbf{x}(t)}{\partial t} = - \mathbf{m}(t),\qquad \pdv{ \mathbf{m}(t)}{ t} = - \gamma \mathbf{m}(t)-\bigtriangledown{\mathbf{F}(\mathbf{x}(t))}.
\end{equation}
Replacing $\mathbf{x}(t)$ with $\mathbf{h}(t)$ and parameterizing $-\bigtriangledown$${\mathbf{F}}$ as a neural network $\mathbf{f}(\mathbf{h}(t), t; \theta_{f})$ yields HBNODEs:
\begin{equation}\label{eq:HBNODE}
    \pdv{ \mathbf{h}(t)}{ t} = - \mathbf{m}(t), \qquad \pdv{ \mathbf{m}(t)}{ t} = - \gamma \mathbf{m}(t)+\mathbf{f}(\mathbf{h}(t),t;\theta_{f}),
\end{equation}
that can be rewritten with initial position $\mathbf{h}(t_0)$ and momentum $\mathbf{m}(t_0) := {d\mathbf{h}}/{dt}(t_0)$ as follows:
\begin{equation}\label{eq:HBNODE-one-shot}
    \frac{\partial^2 \mathbf{h}(t)}{\partial t^2} + \gamma \frac{\partial \mathbf{h}(t)}{\partial t} = \mathbf{f}(\mathbf{h}(t), t; \theta_{f}).
\end{equation}
%
%
HBNODE (Eq.~\eqref{eq:HBNODE-one-shot}) requires fewer computations in solving and is well-structured; thus, it also alleviates vanishing gradients.
However, the momentum acceleration often increases too fast, and this leads to the blow-up issue.


\subsection{\solution{}: \underline{Ada}ptive \underline{M}oment Estimation \underline{NODEs}}

Adam~\citep{kingma2014adam} is an optimization method that combines the existing momentum and RMSProp methods. 
It gives inertia to the progressing speed and has an adaptive learning rate according to the amount of change in the curved surface of the recent path.
%
Following the procedure similar to \S\ref{subsec:heavy-ball}, we derive \solution{} from the update rule of Adam:
\begin{equation}\begin{split}\label{eq:Adam}
\left\{\begin{matrix*}[l]
 &  \mathbf{x}_{k+1} &= \mathbf{x}_{k} - s{\mathbf{m}_{k}}/{\sqrt{\mathbf{v}_{k}+\epsilon }}, \\
 &  \mathbf{m}_{k+1} &= \alpha \mathbf{m}_{k} + (1-\alpha) \bigtriangledown{\mathbf{F}(\mathbf{x}_{k+1})}, \\
 &  \mathbf{v}_{k+1} &= \beta \mathbf{v}_{k} + (1-\beta) \bigtriangledown{\mathbf{F}(\mathbf{x}_{k+1})}^{2}.
\end{matrix*}
\right.
\end{split}
\end{equation}

where $\mathbf{m}$ denotes the momentum term and $\mathbf{v}$ denotes the adaptive step term. Converting this rule into a form of ODE gives:
%
\begin{equation}\begin{split}\label{eq:AdamODE}
\left\{\begin{matrix*}[l]
 & \pdv{ \mathbf{x}(t)}{ t} &= - {\mathbf{m}(t)}/{\sqrt{\mathbf{v}(t)+\epsilon }}, \\
 & \pdv{ \mathbf{m}(t)}{ t} &= (1-\alpha) (\bigtriangledown{\mathbf{F}(\mathbf{x}(t))} - \mathbf{m}(t)), \\
 & \pdv{ \mathbf{v}(t)}{ t} &= (1-\beta) (\bigtriangledown{\mathbf{F}(\mathbf{x}(t))}^{2} - \mathbf{v}(t)). \\
\end{matrix*}
\right.
\end{split}\end{equation}
As has been done in deriving HBNODEs (Eq.~\eqref{eq:HBNODE}), replacing $\mathbf{x}(t)$ with $\mathbf{h}(t)$ and parameterizing $-\bigtriangledown{\mathbf{F}}$ as a neural network $\mathbf{f}(\mathbf{h}(t), t, \theta_{f})$ yields \solution{} such that:
%
\begin{equation}\begin{split}\label{eq:AdamNODE}
\left\{\begin{matrix*}[l]
 &  \frac{\partial \mathbf{h}(t)}{\partial t} &= - {\mathbf{m}(t)}/{\sqrt{\mathbf{v}(t)+\epsilon }}, \\
 &  \frac{\partial \mathbf{m}(t)}{\partial t} &= (1-\alpha) (-\mathbf{f}(\mathbf{h}(t),t;\theta_{f}) - \mathbf{m}(t)), \\
 &  \frac{\partial \mathbf{v}(t)}{\partial t} &= (1-\beta) ({\mathbf{f}(\mathbf{h}(t),t;\theta_{f})}^{2} - \mathbf{v}(t)). \\
\end{matrix*}\right.
\end{split}\end{equation}
%

\subsection{Adjoint Equation of \solution{}}
\label{subsec:adjoint}

If we denote the loss, which measures the discrepancy between the prediction $\mathbf h(t_1)$ and the ground truth as $\mathbf{L}$, the backward pass to update the model parameters $\theta_f$ can be represented as systems of ODEs:
%
\begin{align}
    \dv{ \mathbf{L}(\mathbf{h}(t_1))}{ t} &= \int_{t_0}^{t_1} \mathbf{a}(t)	\pdv{\mathbf{f}(\mathbf{h}(t), t; \theta_{f})}{ \theta} \mathrm d t ,\\
    \dv{ \mathbf{a}(t)}{ t} &= -\mathbf{a}(t) \pdv{ \mathbf{f}(\mathbf{h}(t), t; \theta_{f})}{ \mathbf{h}},
\end{align}
where $\mathbf{a}(t) = \pdv{ \mathbf{L}}{ \mathbf{h}(t)}$ is the adjoint state.
Satisfying the adjoint state is important in reducing the computations as it allows us to compute the backward pass by solving the reverse-mode IVP from the final state towards the initial state.
The adjoint equations of \solution{} are written as follows (we skip the full derivation process due to the space limitation):
\begin{equation}\begin{split}\label{eq:adjointAdamNODE}
\left\{\begin{matrix*}[l]
 & \pdv{\mathbf{a_{h}}(t)}{ t} &= - (-\pdv{ \mathbf{f}(\mathbf{h}(t),t;\theta_{f})}{ \mathbf{h}}\mathbf{a}_{m}(t) + \pdv{ \mathbf{f}(\mathbf{h}(t),t;\theta_{f})}{ \mathbf{h}}\mathbf{a}_{v}(t)),\\
 &  \pdv{ \mathbf{a_{m}}(t)}{ t} &= - (-\frac{1}{\sqrt{v(t)+\epsilon}}\mathbf{a_{h}}(t) -(1-\alpha)\mathbf{a_{m}}(t)),\\
 &  \pdv{ \mathbf{a_{v}}(t)}{ t} &= - (\frac{m}{2(v(t)+\epsilon)\sqrt{v(t)+\epsilon}}\mathbf{a_{h}}(t) - (1-\beta) \mathbf{a_{v}}(t)).
\end{matrix*}\right.
\end{split}\end{equation}
%
%
We solve the adjoint equations using the computationally efficient technique proposed by~\citet{dupont2019augmented}.
Detailed proofs can be found in Appendix~\ref{sec:adjoint_adamnode}.


%
\section{Experimental Results}
\label{sec:results}

We now evaluate the accuracy, efficiency, and stability of \solution{} using the following metrics:

\topic{Metrics.}
We use the \emph{classification accuracy} (henceforth, we call \emph{accuracy}) on the train-set to measure the generalization performance.
To measure the computational efficiency of \solution{}, we define the test accuracy per the number of function evaluations (NFE), \emph{efficacy} in short, in the forward and backward passes.
Efficacy shows the performance advantage of a model per the unit of computations (NFE).
Following the prior work~\cite{xia2021heavy}, we measure the stability of \solution{} by computing the $\ell_2$-norm of $\mathbf{h}(t)$ over time.
The smaller the $|\mathbf{h}(t)|_{\ell_2}$ is, the more stable a model is.

\topic{Datasets and baselines.}
We consider five existing NODEs as our baselines: NODE~\citep{chen2018neural}, ANODE~\citep{dupont2019augmented}, SONODE~\citep{norcliffe2020second}, HBNODE~\citep{xia2021heavy}, and GHBNODE~\citep{xia2021heavy}.
To compare the accuracy and efficacy, we train them on two image classification benchmarks: MNIST~\cite{lecun2010mnist} and CIFAR-10~\citep{CIFAR}.
We also experiment on the Silverbox tasks~\cite{norcliffe2020second} to compare the stability of the training process.
In Silverbox, the rapid increase of $|\mathbf{h}(t)|_{\ell_2}$ leads to the finite-time \emph{blow-up} and makes it difficult to train models.
For a fair comparison, we made the number of parameters of all our models close to each other.


To train those models, we use the Adam optimizer~\citep{kingma2014adam}.
In CIFAR10, we set the learning rate to 0.001 and batch size to 64.
We use the learning rate of 0.0001 and the batch size of 32 in MNIST.
We use Dormand-Prince-45 as the numerical ODE solver.
For HBNODE and GHBNODE, we set the damping parameter $\gamma$ to $sigmoid(\theta)$, where $\theta$ is a trainable and initialized to -3.
We use a 3-layer convolutional neural network to parameterize $f(\mathbf{h}(t), t;\theta_{f})$.
We use Python v3.7 and TorchDiffEq\footnote{https://github.com/rtqichen/torchdiffeq}.
We run all our experiments on a machine equipped with 3 Nvidia RTX A6000 GPUs.

\begin{figure}[t]
    \centering
    \includegraphics[width=0.24\textwidth]{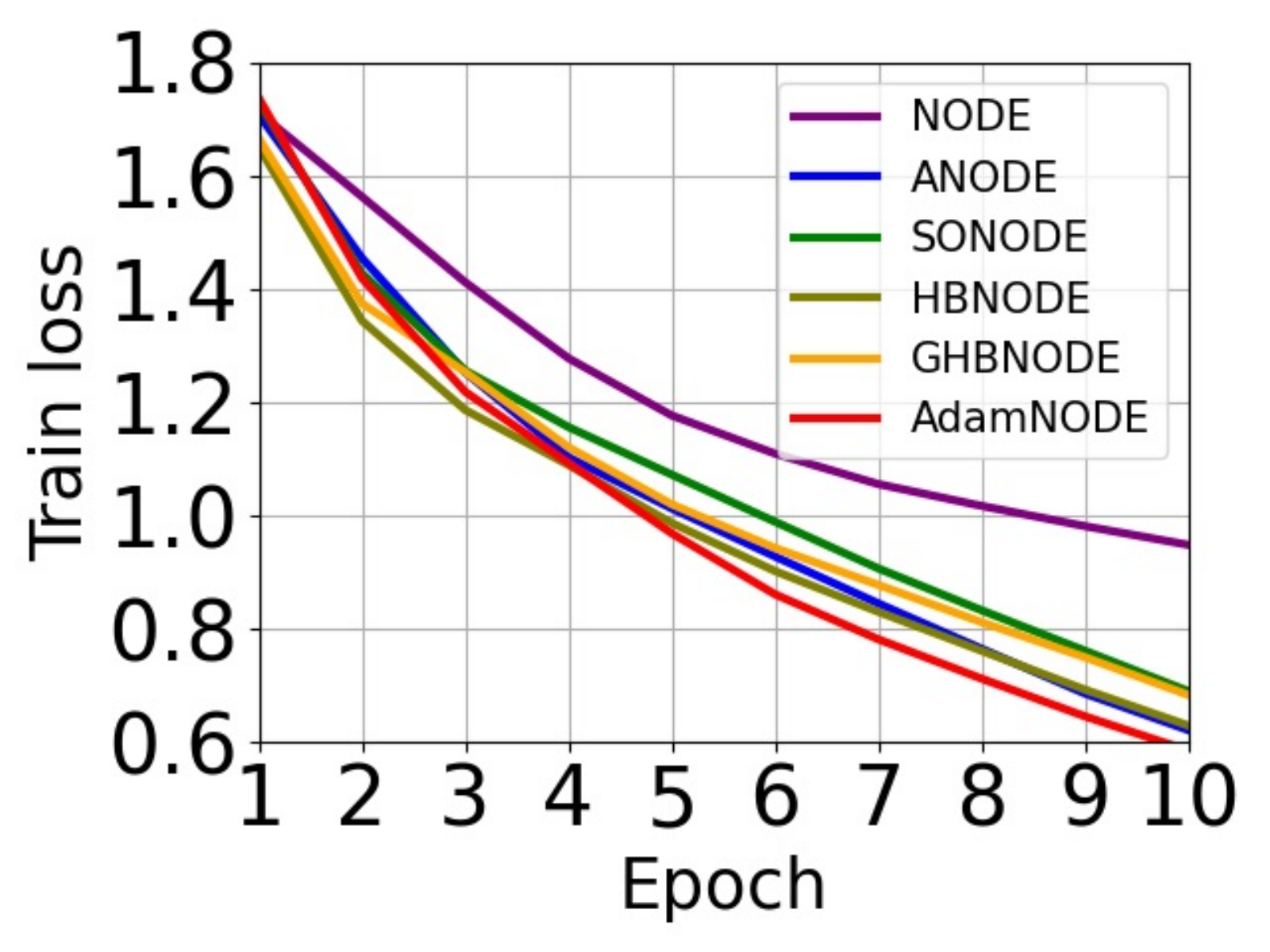}
    \includegraphics[width=0.24\textwidth]{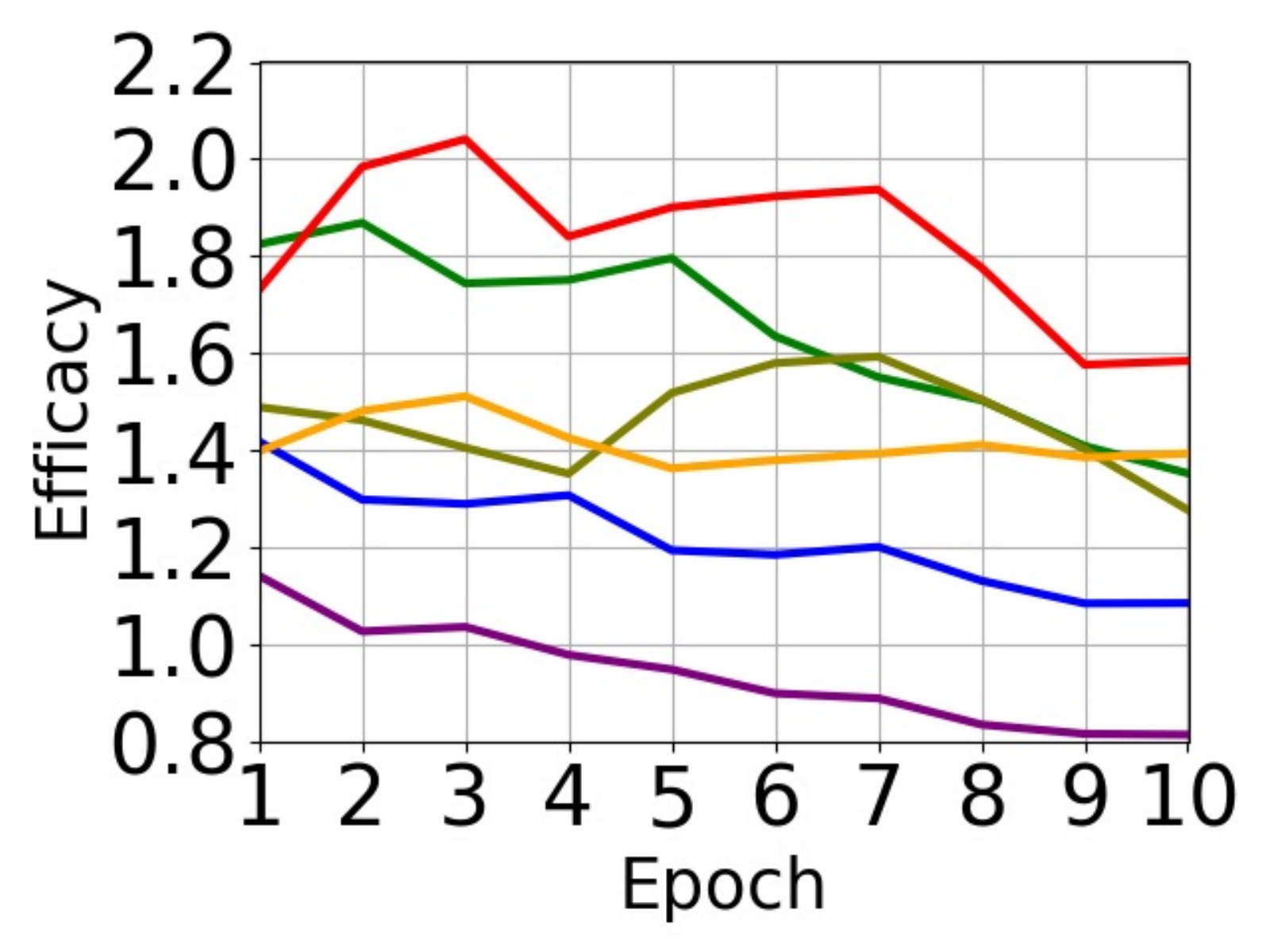}
    \includegraphics[width=0.24\textwidth]{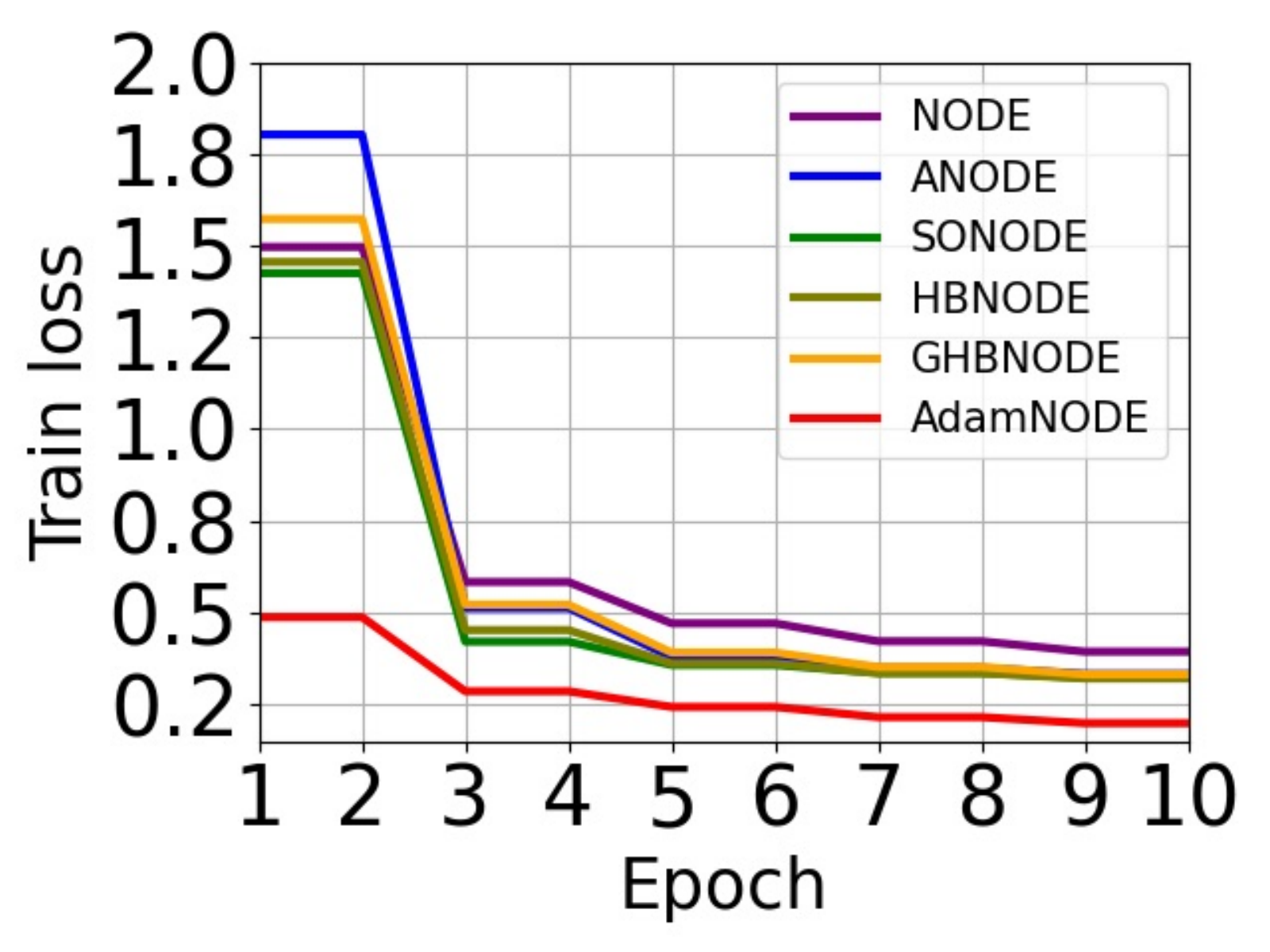}
    \includegraphics[width=0.24\textwidth]{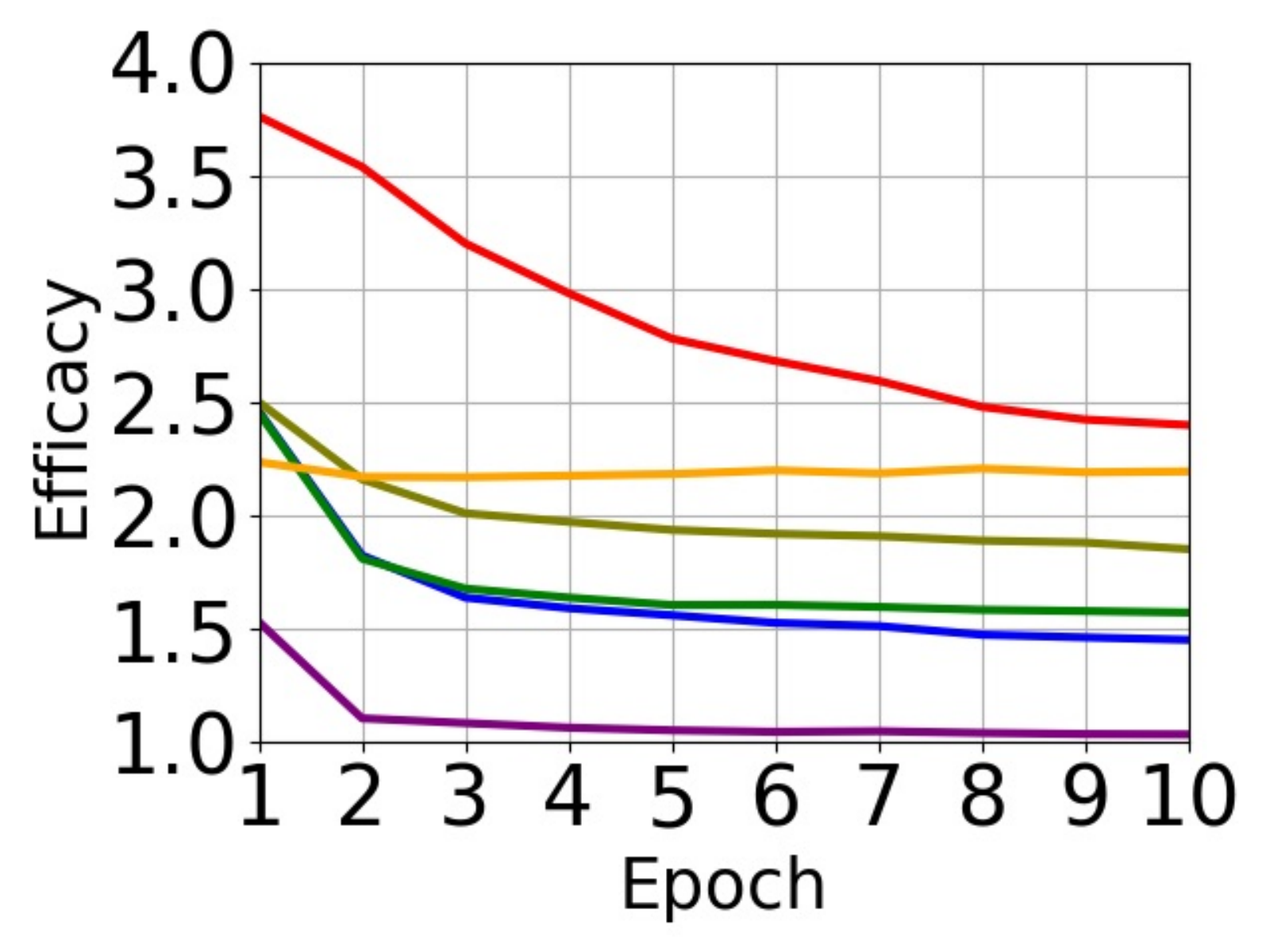}
    \vspace{-.5em}
    \caption{\textbf{Contrasting \solution{} and the five baseline models trained on CIFAR-10 and MNIST.} The left two figures show our CIFAR10 results, and the results from MNIST are shown in the right two figures. We show the training loss and efficacy (\textit{i.e.}, the accuracy per NFE) over 10 epochs. In both CIFAR-10 and MNIST, \solution{} achieve the lowest training loss and the highest efficacy. We also compare the training time and the memory efficiency of \solution{} with the baseline models in Appendix~\ref{sec:trainingtime} and~\ref{sec:memory}, respectively.}
    \vspace{-.5em}
    \label{fig:image_classification}
\end{figure}


\begin{wrapfigure}{r}{5.6cm}
    \centering
    \vspace{-1.2em}
    \includegraphics[width=\linewidth]{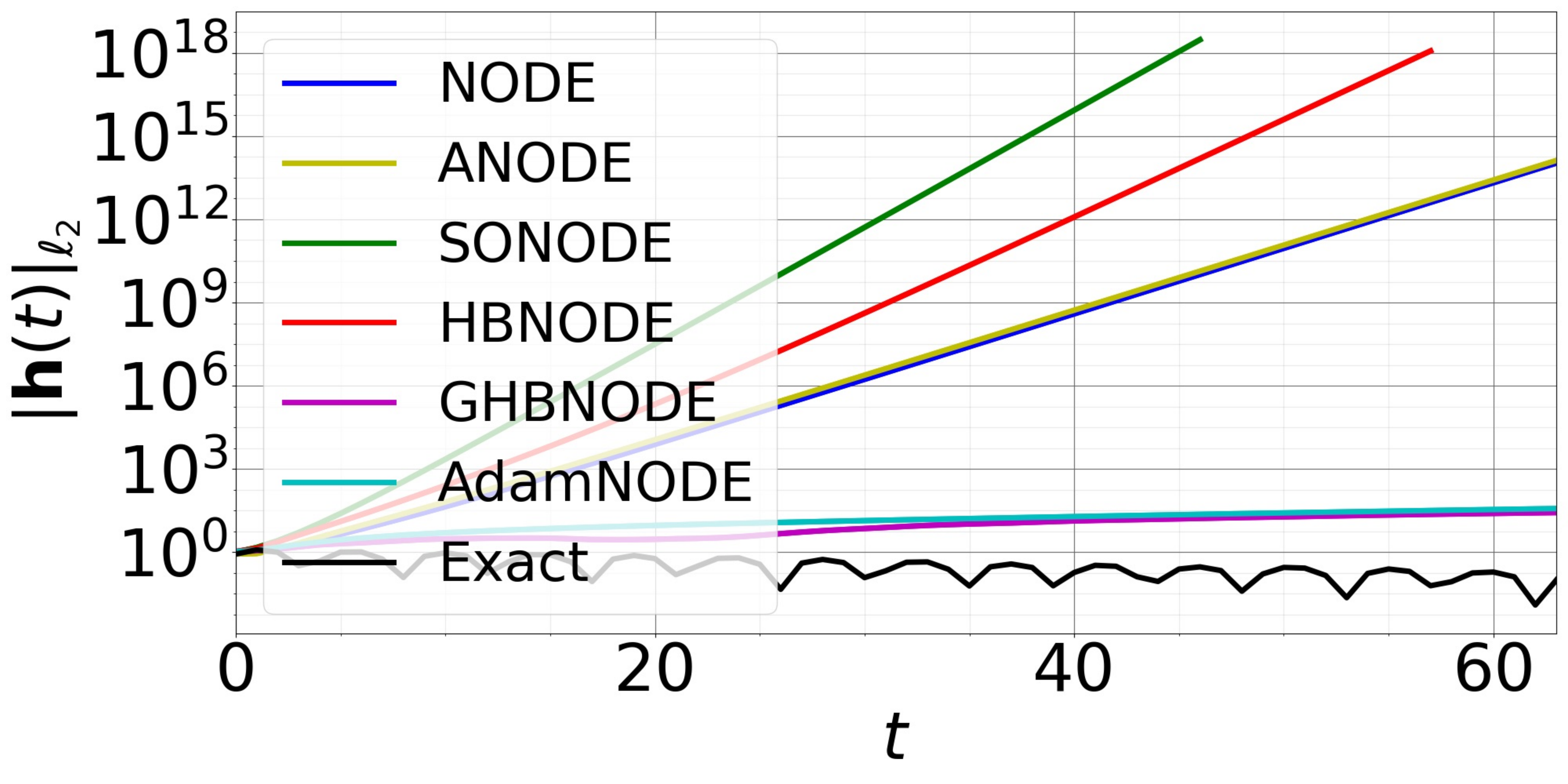}
    \vspace{-2.em}
    \caption{\textbf{Contrasting $\mathbf{h}(t)$ for six models.} $|\mathbf{h}(t)|_{\ell_2}$ in HBNODE increases much faster than \solution{} as $t$ increases.}
    \label{fig:blow_up}
    \vspace{-1.5em}
\end{wrapfigure}
\topic{Image classification results.}
Figure~\ref{fig:image_classification} illustrates our MNIST and CIFAR-10 results.
We first highlight that \solution{} achieve the lower training loss and higher efficacy over the baseline models. 
In particular, the efficacy of \solution{} is 20\% higher than the momentum-based HBNODEs and 80\% higher than the vanilla NODEs.
We observe the consistent results in MNIST. AdamNODE's accuracy is 0.95\% point higher than the best existing model --- we omit the accuracy table for space reasons.
%



\topic{The Silverbox task results.}
We also show that AdamNODE stabilizes the training process.
The results are shown in Figure~\ref{fig:blow_up}.
We train the five baseline models and AdamNODE on the Silverbox task and plot the changes in $|\mathbf{h}(t)|_{\ell_2}$ as $t$ increases.
We find that, at $t$=64, AdamNODE has $10^{18}\times$ smaller $|\mathbf{h}(t)|_{\ell_2}$ value than that of HBNODE.
We also show that, compared to the $|\mathbf{h}(t)|_{\ell_2}$ value of NODE at $t$=64, AdamNODE has $10^{12}\times$ smaller one.
GHBNODE shows the $|\mathbf{h}(t)|_{\ell_2}$ value similar to that of AdamNODE.
Note that AdamNODE achieves a similar stability without using any bounded activation function that requires for GHBNODEs.




%
\section{Conclusions and Future Work}
\label{sec:conclusion}

This work proposes \solution{} to improve the performance, efficiency, and stability of the training and inference of NODEs further.
We show that \solution{} achieve the lowest training loss and improve the training and inference efficacy by 20--80\% over existing baseline NODEs.
We also demonstrate that \solution{} mitigate the blow-up issue without taking the ad-hoc solutions the classical momentum-based neural ODEs (\textit{e.g.}, HBNODEs or GHBNODEs) use.

\topic{What's Next?}
The result suggests that future work is needed for adapting advanced algorithms proposed in the optimization community to NODEs.
As shown in Figures~\ref{fig:image_classification} and~\ref{fig:blow_up}, we may achieve a better utility in conjunction with the ease of training.
One can also explore the use of adaptive moment estimation in generative modeling or normalizing flows.

%

%
\newpage



%
\nocite{langley00}
{
    \bibliography{thispaper}
    \bibliographystyle{icml2022}
}

\appendix
\onecolumn
\section{The Adjoint Equation for AdamNODEs}
\label{sec:adjoint_adamnode}


We describe the details of deriving the adjoint equations for AdamNODEs as follows:
\begin{equation}\begin{split}\label{eq:AdamNODE_appendix}
\left\{\begin{matrix*}[l]
 &  \frac{\partial \mathbf{h}(t)}{\partial t} &= - {\mathbf{m}(t)}/{\sqrt{\mathbf{v}(t)+\epsilon }}, \\
 &  \frac{\partial \mathbf{m}(t)}{\partial t} &= (1-\alpha) (-\mathbf{f}(\mathbf{h}(t),t;\theta_{f}) - \mathbf{m}(t)), \\
 &  \frac{\partial \mathbf{v}(t)}{\partial t} &= (1-\beta) ({\mathbf{f}(\mathbf{h}(t),t;\theta_{f})}^{2} - \mathbf{v}(t)). \\
\end{matrix*}\right.
\end{split}\end{equation}
Denote that the initial state $t\!=\!t_0$ and final state $t\!=\!T$ as 
\begin{equation}
\begin{bmatrix}
\mathbf{h} \\ \mathbf{m} \\ \mathbf{v} \\
\end{bmatrix}\\
(t_{0}) = \\
\begin{bmatrix}
\mathbf{h}_{t_{0}}\\ \mathbf{m}_{t_{0}}\\ \mathbf{v}_{t_{0}} \\
\end{bmatrix}, \\
\begin{bmatrix}
\mathbf{h} \\ \mathbf{m} \\ \mathbf{v} \\
\end{bmatrix}\\
(T) = \\
\begin{bmatrix}
\mathbf{h}_{T}\\ \mathbf{m}_{T}\\ \mathbf{v}_{T} \\
\end{bmatrix} \\
= z \\
.
\end{equation}

Using the proof from~\citet{xia2021heavy}, we have the adjoint equations as the following form:
\begin{equation}
\frac{\partial \mathbf{A}(t)}{\partial t} = 
-A(t) \\
\begin{bmatrix}
0 & \frac{-1}{\sqrt{v(t)+\epsilon}} & \frac{m}{2(v(t)+\epsilon)\sqrt{v(t)+\epsilon}} \\
-\pdv{ \mathbf{f}(\mathbf{h}(t),t;\theta_{f})}{ \mathbf{h}} & (1-\alpha)I & 0 \\
\pdv{ \mathbf{f}(\mathbf{h}(t),t;\theta_{f})}{ \mathbf{h}} & 0 & (1-\beta)I \\
\end{bmatrix}, 
\\
\mathbf{A}(T)= 
-\mathbf{I},  
\\
\mathbf{a}(t) = 
\frac{\partial L}{\partial z}\mathbf{A}(t).
\end{equation}

Suppose $\mathbf{a} = \big[ \mathbf{a}_{h}\; \mathbf{a}_{m}\; \mathbf{a}_{v} \big]$, 
the above equations become:
\begin{equation}\label{eqn:eq14}
\frac{\partial \begin{bmatrix*} \mathbf{a}_{h}&\mathbf{a}_{m}&\mathbf{a}_{v} \end{bmatrix*}}{\partial t} = 
-\begin{bmatrix*} \mathbf{a}_{h}&\mathbf{a}_{m}&\mathbf{a}_{v} \end{bmatrix*} \\
\begin{bmatrix}
0 & \frac{-1}{\sqrt{v(t)+\epsilon}} & \frac{m}{2(v(t)+\epsilon)\sqrt{v(t)+\epsilon}} \\
-\pdv{ \mathbf{f}(\mathbf{h}(t),t;\theta_{f})}{ \mathbf{h}} & (1-\alpha)I & 0 \\
\pdv{ \mathbf{f}(\mathbf{h}(t),t;\theta_{f})}{ \mathbf{h}} & 0 & (1-\beta)I \\
\end{bmatrix},
\end{equation}
We now simplify Eq.~\ref{eqn:eq14} as follows:
\begin{equation}\begin{split}
\left\{\begin{matrix*}[l]
 & \pdv{\mathbf{a_{h}}(t)}{ t} &= - (-\pdv{ \mathbf{f}(\mathbf{h}(t),t;\theta_{f})}{ \mathbf{h}}\mathbf{a}_{m}(t) + \pdv{ \mathbf{f}(\mathbf{h}(t),t;\theta_{f})}{ \mathbf{h}}\mathbf{a}_{v}(t)),\\
 &  \pdv{ \mathbf{a_{m}}(t)}{ t} &= - (-\frac{1}{\sqrt{v(t)+\epsilon}}\mathbf{a_{h}}(t) -(1-\alpha)\mathbf{a_{m}}(t)),\\
 &  \pdv{ \mathbf{a_{v}}(t)}{ t} &= - (\frac{m}{2(v(t)+\epsilon)\sqrt{v(t)+\epsilon}}\mathbf{a_{h}}(t) - (1-\beta) \mathbf{a_{v}}(t)).
\end{matrix*}\right.
\end{split}\end{equation}

\section{Training Time} 
\label{sec:trainingtime}

%
\begin{figure}[h]
\begin{center}
\includegraphics[width=0.27\linewidth]{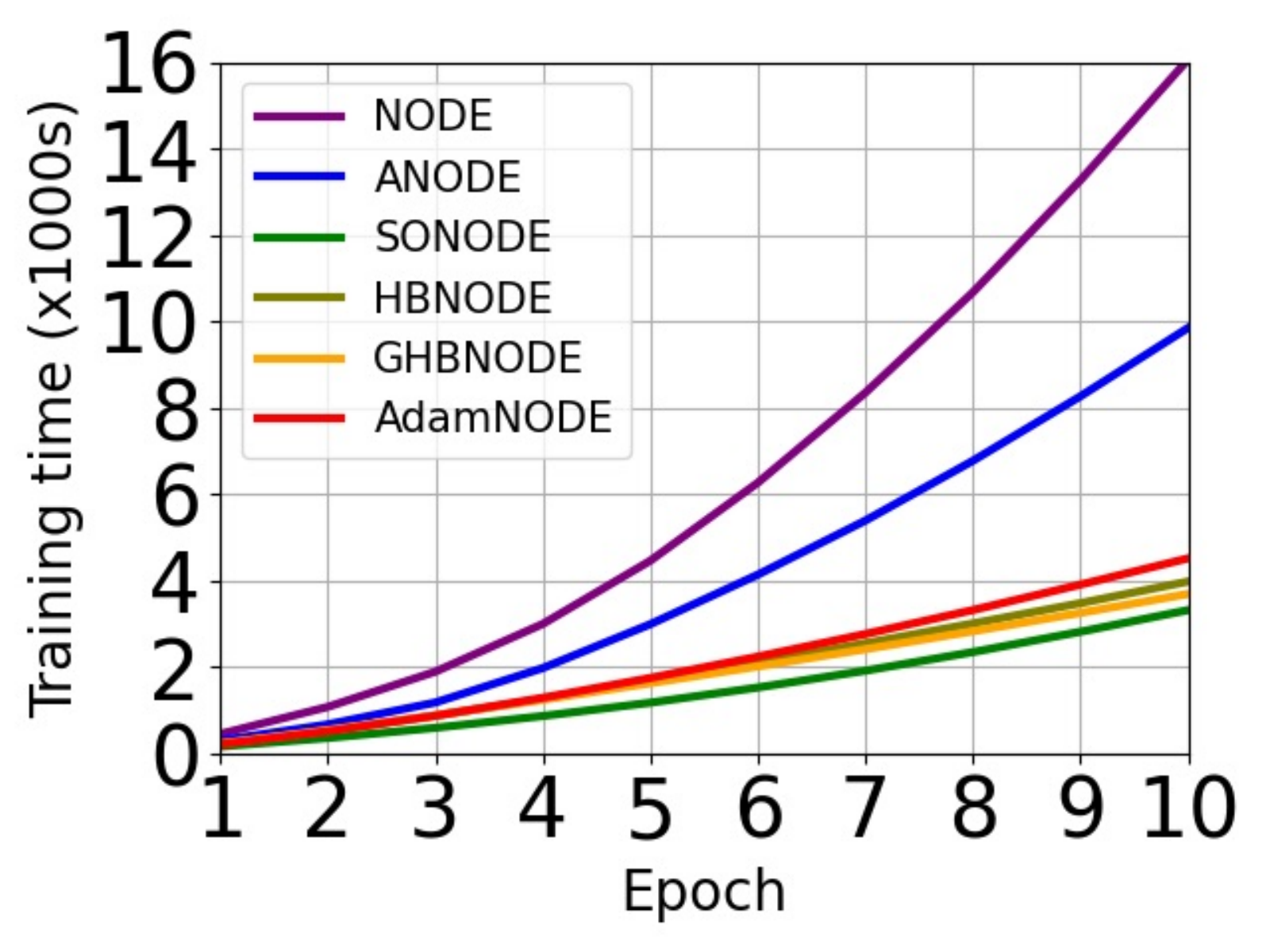}
\includegraphics[width=0.27\linewidth]{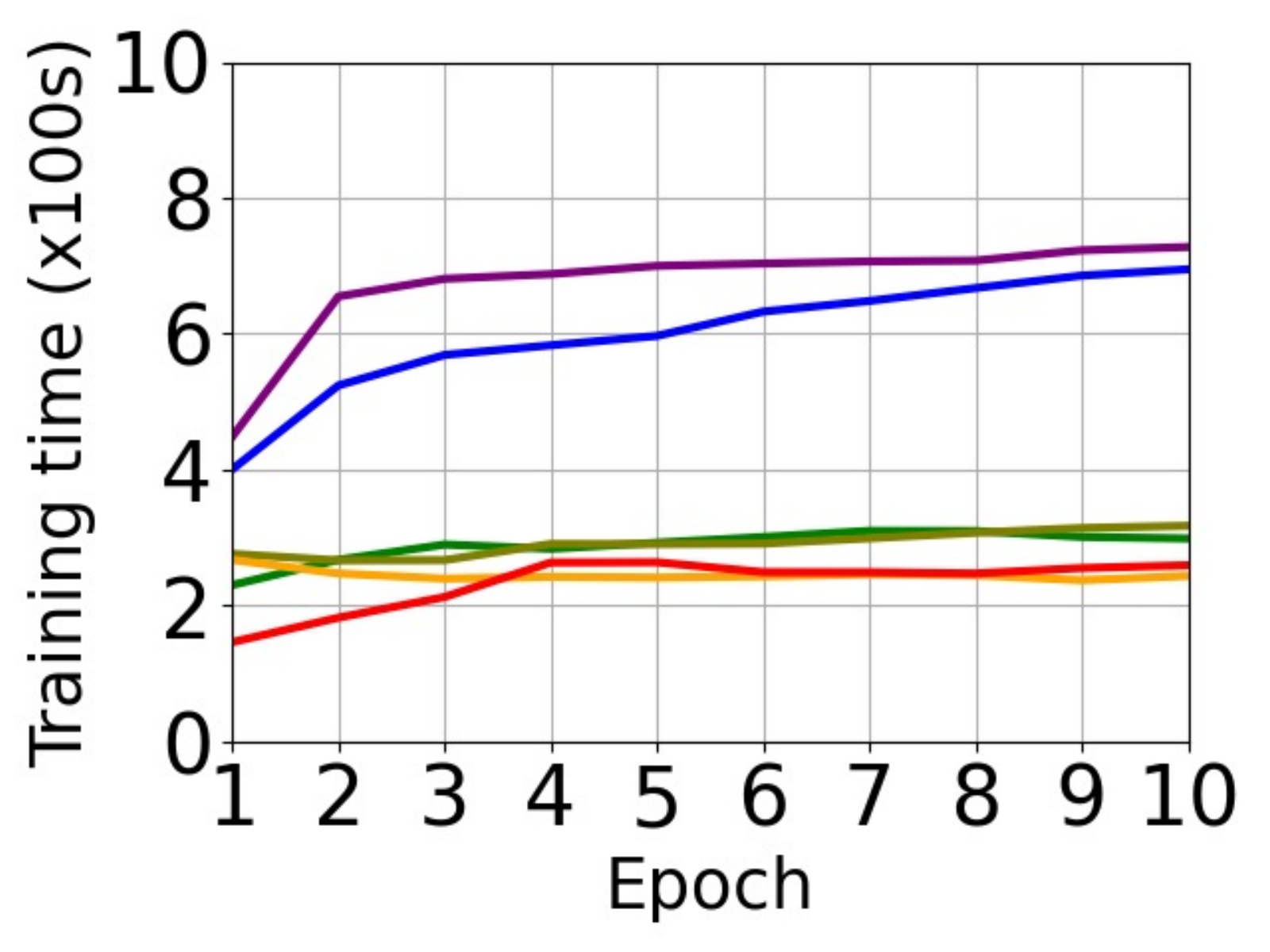}
\end{center}
\vspace{-1.6em}
\caption{\textbf{Comparison of the training time for different models.} We compare the training time of five baseline NODEs with AdamNODEs. We measure the wall-time difference between the start and the end of training. We show the CIFAR10 (left) and MNIST (right) results. We find that AdamNODEs achieve the training time similar to the momentum-based HBNODEs and GHBNODEs.}
\label{fig:trainingtime}
\end{figure}
%
 
%
%
%
Here, we examine whether AdamNODEs increase the training time of the momentum-based models, such as HBNODEs or GHBNODEs.
Due to the adaptive moment estimation, we hypothesize that AdamNODEs may require a longer training time than the momentum-based NODEs.
Hence, we compare the training time for the NODEs we use in our experiments.
The results are shown in Figure~\ref{fig:trainingtime}.
We observe that AdamNODEs do not increase the training time of HBNODEs or GHBNODEs.
In both CIFAR10 and MNIST, AdamNODEs' training time is the same as that of the momentum-based NODEs.

\newpage

\section{Memory Efficiency}
\label{sec:memory}

\begin{table}[ht]
\begin{center}
\begin{tabular}{c|rrrrrr}
\toprule
 & \multicolumn{1}{c}{\textbf{NODE}} & \multicolumn{1}{c}{\textbf{ANODE}} & \multicolumn{1}{c}{\textbf{SONODE}} & \multicolumn{1}{c}{\textbf{HBNODE}} & \multicolumn{1}{c}{\textbf{GHBNODE}} & \multicolumn{1}{c}{\textbf{AdamNODE}} \\ \midrule \midrule
\textbf{CIFAR10} & 2469 MiB                  &       2485 MiB                     &                 2863 MiB            &                    2891 MiB         &                      2889 MiB        & 
3531 MiB \\ 
\\ \bottomrule
\end{tabular}
\end{center}
\vspace{-0.9em}
\caption{\textbf{Comparison of memory efficiency for different models.} We compare the memory efficiency of five baseline NODEs with our AdamNODEs. We measure the memory footprints in MiB. We find that AdamNODEs require slightly more memory than the baselines.}
\label{fig:memory}
\end{table}

We compare the memory footprints of AdamNODEs with five baselines.
Table~\ref{fig:memory} shows our results.
We confirmed that the memory footprints of NODEs and ANODEs are the same.
%
The memory footprint is larger in SONODEs and (G)HBNODEs than ANODEs as there are additional variables (e.g., velocity).
AdamNODEs show the increased memory footprint as there are more additional variables (e.g., velocity or variables for adaptive moment estimation).



\end{document}